\documentclass{article}
\usepackage{spconf,amsmath,graphicx}
\usepackage[dvipsnames]{xcolor}
\usepackage{booktabs}
\usepackage{multirow}

\usepackage[colorlinks=True,
            linkcolor=red,
            filecolor=red,
            anchorcolor=red,
            citecolor=blue]{hyperref}

\usepackage{url}

\title{SSGD: A smartphone screen glass dataset for defect detection}
\name{Haonan Han$^\dag$ \thanks{$\dag$ Equal contribution, $\ddag$ Corresponding author}, Rui Yang$^\dag$, Shuyan Li, Runze Hu, Xiu Li$^\ddag$}
\address{Tsinghua Shenzhen International Graduate School, Tsinghua University, China}
\begin{document}
%
\maketitle
\begin{abstract}
Interactive devices with touch screen have become commonly used in various aspects of daily life, which raises the demand for high production quality of touch screen glass.
While it is desirable to develop effective defect detection technologies to optimize the automatic touch screen production lines, the development of these technologies suffers from the lack of publicly available datasets.
To address this issue, we in this paper propose a dedicated touch screen glass defect dataset which includes seven types of defects and consists of 2504 images captured in various scenarios.
All data are captured with professional acquisition equipment on the fixed workstation.
Additionally, we benchmark the CNN- and Transformer-based object detection frameworks on the proposed dataset to demonstrate the challenges of defect detection on high-resolution images.
Dataset and related code will be available at \url{https://github.com/VincentHancoder/SSGD}.
\end{abstract}
%

\begin{figure*}[t]
\begin{minipage}[b]{1.0\linewidth}
  \centering
  \centerline{\includegraphics[width=1.0\linewidth]{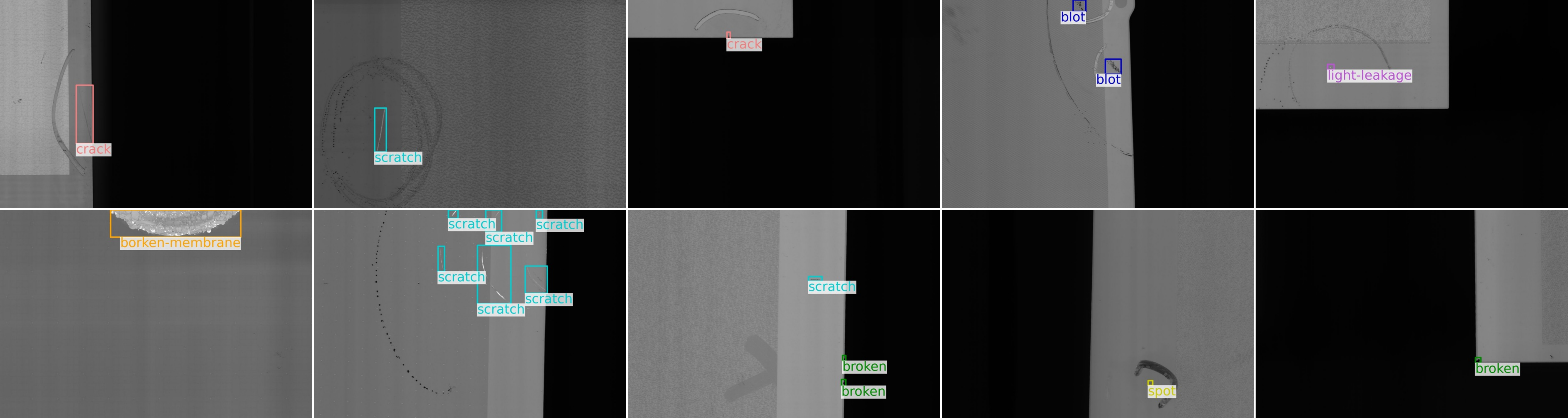}}
  \centerline{(a) Samples of Part I.}\medskip
\end{minipage}
\hfill
\begin{minipage}[b]{1.0\linewidth}
  \centering
  \centerline{\includegraphics[width=1.0\linewidth]{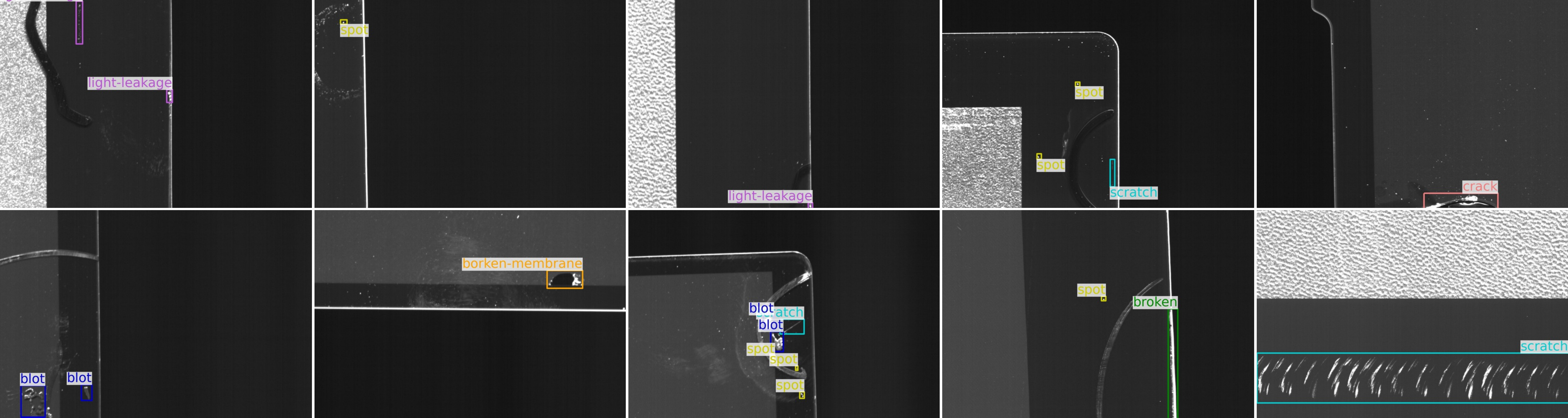}}
  \centerline{(b) Samples of Part II.}
\end{minipage}
\caption{Sample visualization for SSGD. (a) Some samples in SSGD Part I. (b) Some samples in SSGD Part II.}
\label{fig:vis_sample}
\end{figure*}

\section{Introduction}
\label{sec:intro}

Nowadays, smart terminals are increasingly crucial to the intelligence process of contemporary society. Almost everyone has at least one smartphone in their hands. As the essential accessories, the production quality of the smartphone screen directly determines the display effect, service life, and user evaluation of smartphones. To this end, almost all screens must pass a production quality inspection before they leave the factory. However, defect detection via human beings only is labor-intensive and inefficient, which is insufficient for the vast market of smartphones. In this case, the use of data-driven computer vision inspection methods can significantly improve efficiency and reduce judgment errors brought on by human factors.

There have been several attempts to apply deep learning techniques to the task of detecting defects in industry. In the metal generic surface defect detection area, Bao et al.~\cite{bao2021triplet} presented a dedicated dataset called NEU-Dataset, and Song et al.~\cite{song2013micro} raised a matching method with a set of data used for defect detection of silicon steel strip micro surface. Similarly, a railway surface defect dataset~\cite{gan2017hierarchical} was made for detection in 2017. Besides metal surface detection tasks, increasing number of industrial products dataset~\cite{pallemulla2021defect, pramerdorfer2015dataset, Deitsch2021} have been widely collected for specific detection scenarios. 
Some works focused on fabric regions~\cite{pallemulla2021defect} have been produced and extended to a challenging competition to encourage contestants to approach higher detection accuracy.
Recently more in-depth, increasing dimensions of the electronics industry production have been using neural network technologies to detect the quality of products and defect distribution. 
Pramerdorfer et al.~\cite{pramerdorfer2015dataset} announced a dataset of Printed circuit board (PCB) which was made to facilitate the computer vision tasks in the challenge of PCB deficiency in producing process.
In 2019, Deitsch et al.~\cite{Deitsch2021} presented a dataset presented a dataset about the solar panels damage situation.

However, to the best of our knowledge, there is no publicly available dataset for defect detection of smartphone screens.~This seriously hinders the application of computer vision technology in the defects inspection of screens.
To solve this problem, we firstly propose an open-source \textbf{S}martphone \textbf{S}creen \textbf{G}lass \textbf{D}ataset, dubbed as \textbf{SSGD}, which contains basically common types of defect occurring on the glass panels.
Specifically, the proposed SSGD is made up of 2504 images and contains seven types of defects commonly existing in the production process. All images are in an uniformed resolution of $1500\times 1000$ pixels.  Figure~\ref{fig:vis_sample} shows some samples of SSGD. 
After the procedure of data collecting and annotating, we conduct extensive experiments based on the general platform to evaluate the performance of popular object detection on SSGD (Sec.~\ref{sec:exp}).

Our main contributions can be summarized as follows:
\begin{itemize}
\item We collect and annotate a dataset for defect detection of smartphone screens, which possesses various annotations categories, relatively high image information quality, and public availability.
\item We benchmark many popular object detectors on the proposed dataset, including CNN- and Transformer-based frameworks.
\end{itemize}

\section{Dataset creation procedure}
\label{sec:format}
In this section, we will introduce SSGD from three aspects:
(1) the process of image capturing, (2) the overall properties of the dataset, and (3) the composition of the dataset. 
\subsection{Data Collection}
\label{ssec:subhead}

Before information acquisition, numerous samples with defects have been collected and selected as origin materials for the purpose of capturing. The background scenario is decorated with the color black to weaken the influence of other visible light in the environment on acquired image information. During the image collection process, smartphone touch screens are placed on a specific capture platform which has been calibrated by a leveler to ensure the correctness of the shooting angle. On the platform, the surrounding region of the screen is the ink area, and the middle is the visible area protected by specific film. The device used to collect images is a line-scan camera which is designed for industry-level information acquisition. After the unified collection and initial data cleaning process, we uniformly change the image resolution into $1500\times 1000$ pixels in order to ensure every training object provides the same size background information. The entire dataset is grouped by different capturing platforms and given specific file numbers according to groups. 
After collection, we employ the labelme\footnote{\url{https://github.com/wkentaro/labelme}} to annotate the seven kinds of defects and the corresponding locations (bounding box format)  on the screens and get an XML file for each image.

\subsection{Dataset Properties}
\label{ssec:subhead}

SSGD consists of 2504 images and includes seven common types of defects in actual smartphone screen process. Aiming to show the properties of SSGD clearly, we illustrate the causality in following perspectives:

\noindent\textbf{Categories}: Seven types of defects including \textit{crack}, \textit{broken}, \textit{spot}, \textit{scratch}, \textit{light-leakage}, \textit{blot}, \textit{broken-membrane}. Some samples are provided in Fig.~\ref{fig:vis_sample}.

\noindent\textbf{Workstation}: Two workstations are used to capture images. SSGD is therefore divided into two parts, called Part I and Part II, respectively.

\noindent\textbf{Workstation Content}: Two workstations captured 1258 and 1246 images (corresponding to Part I and Part II).

\noindent\textbf{Part I Content}: the possession situation of each type of defects is that \textit{crack}: 988, \textit{broken}: 304, \textit{spot}: 175, \textit{scratch}: 99, \textit{light-leakage}: 63, \textit{blot}: 18, \textit{broken-membrane}: 10.

\noindent\textbf{Part II Content}: the possession situation of each type of defects is that \textit{crack}: 787, \textit{broken}: 756, \textit{spot}: 467, \textit{scratch}: 163, \textit{light-leakage}: 60, \textit{blot}: 13, \textit{broken-membrane}: 11.
\subsection{Dataset Distribution}
\label{ssec:subhead}

\begin{figure}[t]
\begin{minipage}[b]{1.0\linewidth}
  \centering
  \centerline{\includegraphics[width=1.0\linewidth]{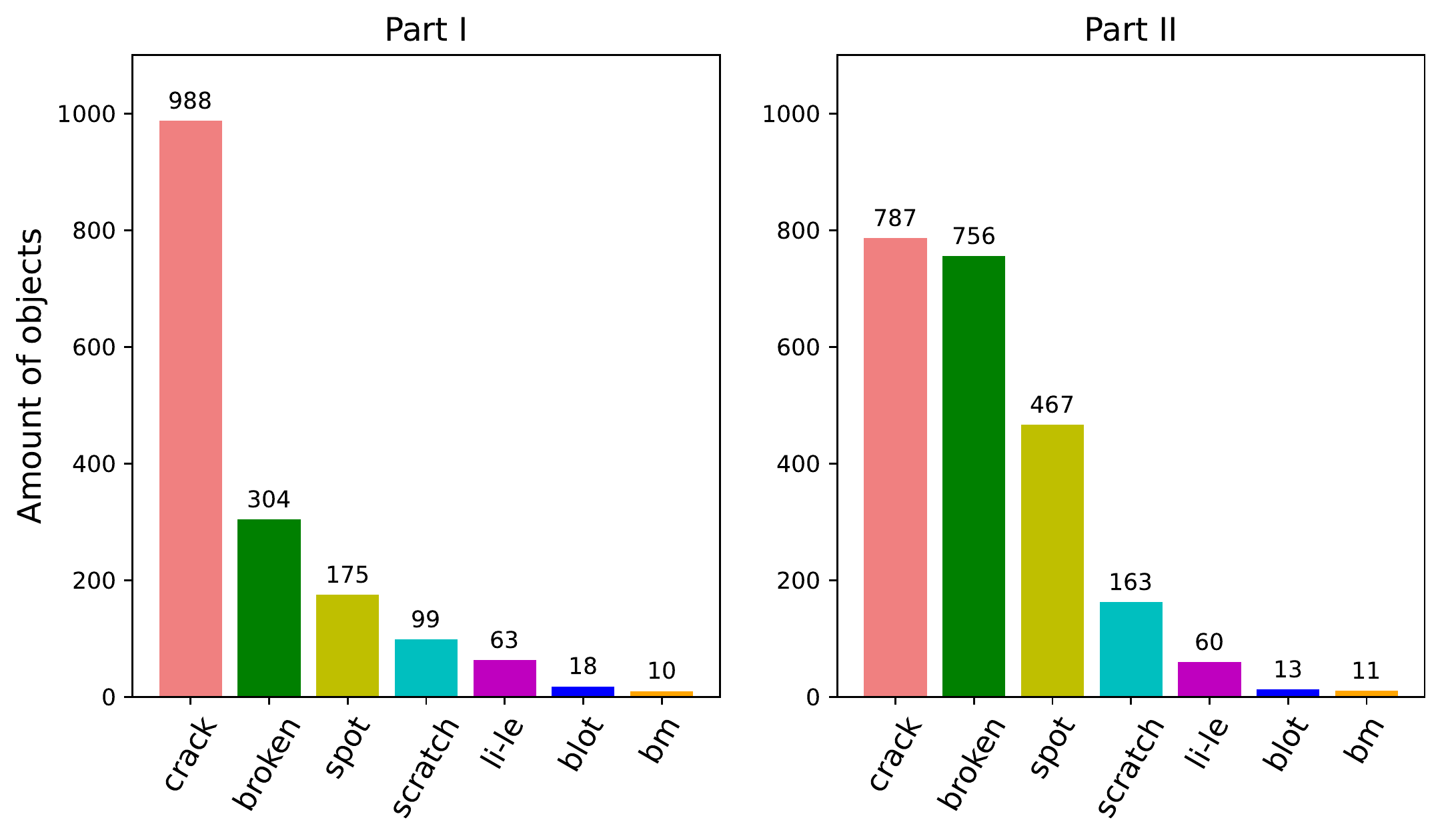}}
  \centerline{(a) Amount of objects}\medskip
\end{minipage}
\hfill
\begin{minipage}[b]{1.0\linewidth}
  \centering
  \centerline{\includegraphics[width=1.0\linewidth]{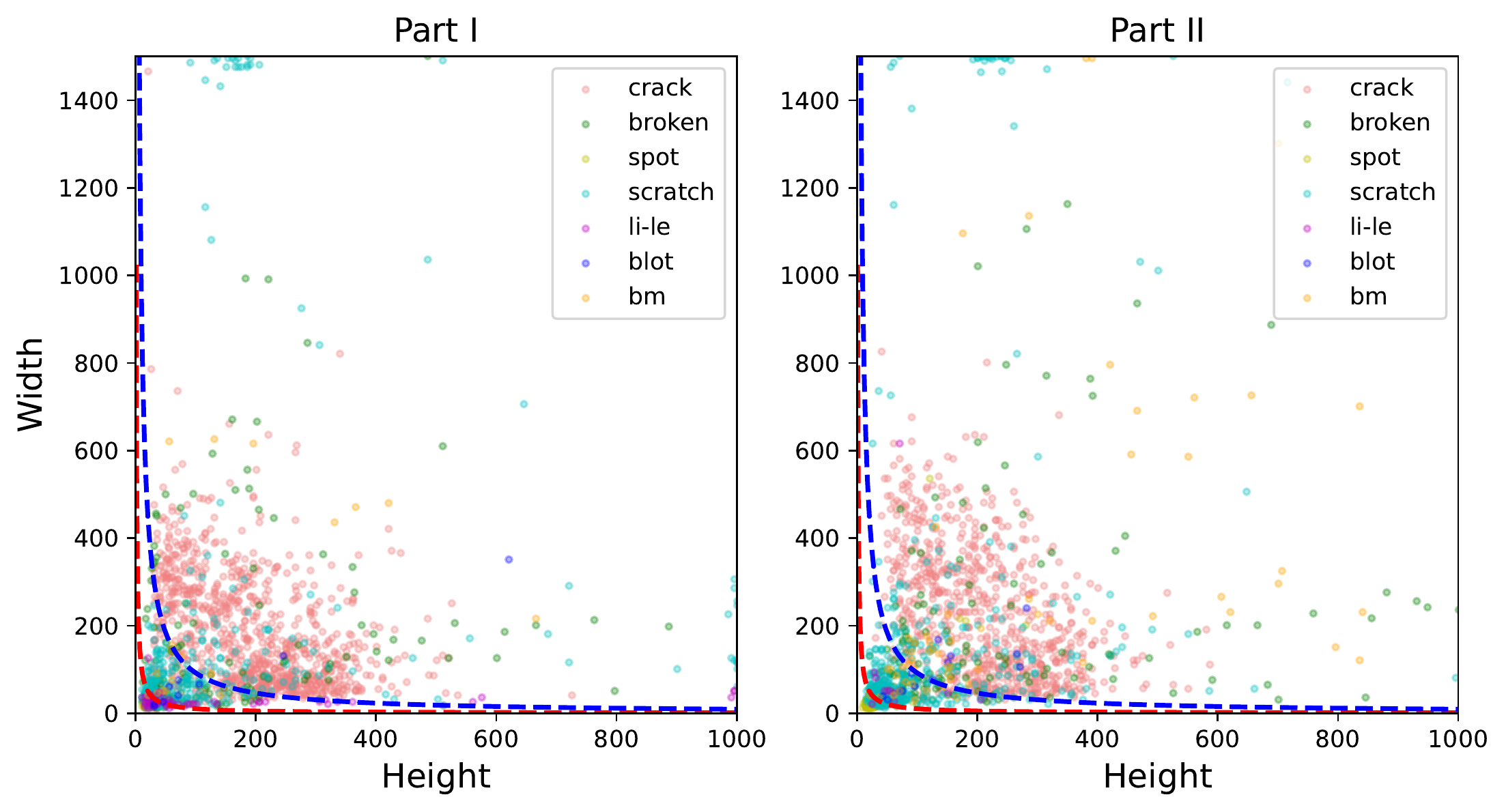}}
  \centerline{(b) The size of objects.}
\end{minipage}
\caption{Object analysis for Part I and Part II of SSGD. 'li-le' denotes light-leakage, and 'bm' refers broken-membrane. In (b), points under the red line are small objects, points between red and blue line are medium objects, and points above the blue line are large objects. Best viewed in color.}
\label{fig:res}
\end{figure}

As shown in Fig.~\ref{fig:res} (a), there are two data bounding box characteristic distribution maps whose horizontal axis and vertical axis represent for the height and the width of each bounding box which are distinguished by different workstations. The position on the map of points in different colors show how their bounding boxes look like. There are two curves in blue and red symbolize thresholds to define small, middle and large target detection object. If the point is lower than red curve, it is a small target covering an area less than $32\times 32$ pixels. Similarly, it is thought as a large target when more distant from origin point than blue curve. And the target which is between blue and red curve will be thought as a middle size detection object.

For the purpose of showing quantity distribution of large, medium and small detection targets, we summarize the following related information that can be corroborated by the relevant information in the Fig.~\ref{fig:res} (a) as well:

\noindent\textbf{Part I}: the amount of different types detection object is  \textit{small}: 241, \textit{middle}: 378, \textit{large}: 1038.

\noindent\textbf{Part II}: the amount of different types detection object is  \textit{small}: 783, \textit{middle}: 441, \textit{large}: 1033.

However, there are some extreme points existing on the map such as the points gathering at the upper left area of the map, which means that bounding boxes represented by those points are at a nearly 7:1 width and length ratio. As the input of network, bounding box in a such extreme shape may influence final convergence direction.

The color intensity shows how gathering a type of defect points are. It is shown that on the both maps (Part I and Part II) category \textit{crack} basically distributed above the blue line while category \textit{scratch} mainly gather at the zone in middle. In other words, we get a pattern that most samples of above two categories are large size and middle size detection object. Obviously, there are still some categories, such as \textit{blot}, that are unable to find the pattern of bounding box size. Such a phenomenon is acceptable as well.

\section{experiment}
\label{sec:exp}

\subsection{Settings}
We benchmark most basic object detection models on the proposed SSGD, including Faster R-CNN~\cite{Faster}, FCOS~\cite{FCOS}, and YOLO series~\cite{YOLOX}. 
To reduce random bias, we use 5-fold cross-validation to measure all models, and results are averaged over the five folds.
In the implementation, most experiments are based on MMDetection~\cite{mmdetection}. YOLOv5 and YOLOX~\cite{YOLOX} follow their official repositories\footnote{\url{https://github.com/ultralytics/yolov5}}\footnote{\url{https://github.com/Megvii-BaseDetection/YOLOX}}.
Before training, we initialize the model with the weight pre-trained on COCO~\cite{COCO} dataset and new layers in the classification head with the Normal scheme. 
During training, for ResNet-50-based~\cite{ResNet} models, we utilize the SGD~\cite{SGD} optimizer and the 2$\times$ (24 epochs) schedules with a global batch size of 16 on 4 GPUs.
We also adopt the multi-scale training, where the short side of input images is randomly resized to [800, 1500] pixels, and the long side is at most 2250 pixels. 
For Transformer-based models~\cite{Swin, PVT, ScalableViT, UniFormer}, we utilize the AdamW~\cite{AdamW} optimizer and the 2$\times$ schedules with a global batch size of 8 on 8 GPUs. The resolution of input images is $1500\times 1000$ pixels. Other settings remain the same as MMDetection.
For YOLO series, the input with a resolution of $1500\times 1000$ pixels is padded to $1500\times 1500$ pixels. We train the model 100 epochs with a global batch size of 16 on 4 GPUs. Other settings remain the same as the original repositories.
During testing, input images maintain their original resolution without any augmentation.

\begin{table*}[t]
\caption{Benchmark general models on SSGD. Besides YOLOv5-m and YOLOX-m, other models take ResNet-50 as the backbone. $\#$Param. refers the total parameters. FLOPs and Frame per seconds (FPS) are tested on single 3090 GPU with the original size.}
\label{tab:my-table}
\resizebox{\textwidth}{!}{
\begin{tabular}{@{}l|ccc|cccccc|cccccc@{}}
\toprule
\multirow{2}{*}{Model} & \multirow{2}{*}{\begin{tabular}[c]{@{}c@{}}$\#$Param.\\ (M)\end{tabular}} & \multirow{2}{*}{\begin{tabular}[c]{@{}c@{}}FLOPs\\ (G)\end{tabular}} & \multirow{2}{*}{\begin{tabular}[c]{@{}c@{}}FPS\\ (img/s)\end{tabular}} & \multicolumn{6}{c|}{Part I}             & \multicolumn{6}{c}{Part II}             \\ \cmidrule(l){5-16} 
                       &                                                                       &                                                                      &                                                                        & AP   & AP$_{50}$ & AP$_{75}$ & AP$_S$  & AP$_M$  & AP$_{L}$  & AP   & AP$_{50}$ & AP$_{75}$ & AP$_S$  & AP$_M$  & AP$_{L}$  \\ \midrule
Faster R-CNN~\cite{Faster}       & 41.2                                                                  & 303.8                                                                & 26.2                                                                   & 19.3 & 41.5 & 14.9 & 15.9 & 23.4 & 25.5 & 23.8 & 46.6 & 21.1 & 20.2 & 22.4 & 25.9 \\
Cascade R-CNN~\cite{Cascade}      & 68.9                                                                  & 331.6                                                                & 21.7                                                                   & 20.9 & 42.3 & 17.4 & 15.0 & 24.0 & 31.2 & 27.3 & 51.1 & 25.6 & 22.3 & 25.4 & \textbf{30.7} \\
RetinaNet~\cite{RetinaNet}          & 36.2                                                                  & 311.2                                                                & 25.0                                                                   & 16.4 & 37.5 & 11.1 & 14.3 & 22.1 & 21.1 & 21.7 & 42.7 & 18.8 & 22.7 & 25.5 & 24.3 \\
FCOS~\cite{FCOS}               & 31.9                                                                  & 296.2                                                                & 28.1                                                                   & 19.4 & 41.9 & 15.7 & 15.7 & 23.4 & 21.6 & 27.1 & 50.7 & 24.8 & 25.9 & 25.6 & 28.5 \\
ATSS~\cite{ATSS}               & 31.9                                                                  & 303.3                                                                & 24.2                                                                   & \textbf{22.3} & \textbf{46.1} & \textbf{18.5} & \textbf{16.6} & \textbf{25.3} & \textbf{26.5} & 27.6 & \textbf{52.8} & \textbf{26.4} & 23.6 & \textbf{27.9} & 26.8 \\
GFL~\cite{GFL}                & 32.1                                                                  & 307.9                                                                & 25.0                                                                   & 19.6 & 43.2 & 15.2 & 15.6 & 24.6 & 23.4 & 27.5 & 50.9 & 25.5 & \textbf{26.2} & 26.7 & 28.4 \\ \midrule
YOLOv5-m$^{\textcolor[rgb]{1,0,0}{2}}$               & 19.9                                                                  & 266.7                                                                & 59.5                                                                   & 16.2 & 38.9 & 11.2 & 13.5 & 22.5 & 18.4 & \textbf{27.8} & 52.4 & 25.0 & 20.4 & 27.2 & 26.8 \\
YOLOX-m~\cite{YOLOX}                & 25.3                                                                  & 405.1                                                                & 36.9                                                                   & 13.4 & 36.2 & 7.8  & 14.7 & 18.3 & 12.7 & 20.7 & 43.5 & 15.7 & 21.3 & 18.0 & 19.0 \\ \bottomrule
\end{tabular}
}
\end{table*}

\begin{table*}[t]
\centering
\caption{Benchmark Transformer-based models on SSGD using Faster R-CNN~\cite{Faster}. $\#$Param. refers the total parameters. FLOPs and Frame per seconds (FPS) are tested on single 3090 GPU with the original size.}
\label{tab:res-t}
\resizebox{\textwidth}{!}{
\begin{tabular}{@{}l|ccc|cccccc|cccccc@{}}
\toprule
\multirow{2}{*}{Model} & \multirow{2}{*}{\begin{tabular}[c]{@{}c@{}}$\#$Param.\\ (M)\end{tabular}} & \multirow{2}{*}{\begin{tabular}[c]{@{}c@{}}FLOPs\\ (G)\end{tabular}} & \multirow{2}{*}{\begin{tabular}[c]{@{}c@{}}FPS\\ (img/s)\end{tabular}} & \multicolumn{6}{c|}{Part I}                  & \multicolumn{6}{c}{Part II}                  \\ \cmidrule(l){5-16} 
                       &                                                                       &                                                                      &                                                                        & AP   & AP$_{50}$  & AP$_{75}$  & AP$_S$   & AP$_M$   & AP$_L$   & AP    & AP$_{50}$  & AP$_{75}$  & AP$_S$  & AP$_M$   & AP$_L$   \\ \midrule
Swin-T~\cite{Swin}                 & 44.8                                                                  & 308.2                                                                & \textbf{18.1}                                                                   & 19.2 & 42.6  & 13.2  & \textbf{15.5}  & 21.5  & \textbf{27.9}  & \textbf{27.0}  & \textbf{52.4}  & \textbf{24.5}  & \textbf{24.8} & \textbf{24.4}  & \textbf{29.7}  \\
PVT-S~\cite{PVT}                  & 41.1                                                                  & 281.3                                                                & 12.3                                                                   & 16.0 & 36.7  & 12.2  & 13.7  & 15.9  & 21.2  & 20.5  & 44.3  & 18.4  & 17.8 & 19.5  & 20.7  \\
ScalableViT-S~\cite{ScalableViT}          & 43.3                                                                  & 297.7                                                                & 10.9                                                                   & \textbf{21.2} & 46.4 & \textbf{15.1} & 14.3 & \textbf{22.1} & 27.3 & 22.8 & 48.3 & 19.3 & 20.7 & 21.5 & 23.4 \\
UniFormer-S$_{h14}$~\cite{UniFormer}       & 38.2                                                                  & 276.4                                                                & 15.8                                                                   & 18.9 & \textbf{45.0}  & 13.7  & 13.7  & 19.9  & 26.7  & 22.2  & 47.3  & 19.1  & 22.1 & 22.4  & 23.0  \\ \bottomrule
\end{tabular}
}
\end{table*}

\subsection{Results}
\noindent \textbf{Results on CNN-based methods.} We evaluate mainstream anchor-based~\cite{Faster, Cascade, RetinaNet, YOLOX} and anchor-free~\cite{FCOS, ATSS, GFL} object detectors on SSGD. Besides YOLOv5 and YOLOX, all other models take ResNet-50 as the backbone. Due to an adaptive training sample selection strategy, ATSS obtains the best performance in all ResNet-50-based models. ATSS outperforms the two-stage Cascade R-CNN by 1.4 AP and 0.3 AP on SSGD Part I and Part II, respectively, while only possessing half the parameters. However, Cascade R-CNN achieves better performance on large objects. 

\noindent \textbf{Results on Transformer-based methods.} We evaluate Swin~\cite{Swin}, PVT~\cite{PVT}, ScalableViT~\cite{ScalableViT}, and UniFormer~\cite{UniFormer} on the proposed SSGD using the Faster R-CNN framework. As reported in Table~\ref{tab:res-t}, ScalableViT-S achieves 21.1 AP on SSGD Part I under a single-scale training strategy, which surpasses most ResNet-50-based methods using the multi-scale training strategy. However, on SSGD Part II, Swin-T obtains better performance (27.0 AP) than other Vision Transformer counterparts. When compared to ResNet-50, Swin-T gets 3.2 AP gains. Nevertheless, Transformer-based models have an obvious disadvantage in the speed that is required in industrial scenarios. Specifically, when input resolution is $1500 \times 1000$ pixels, Swin-T-based Faster R-CNN can only process 18 images per second, but ResNet-50-based ones can process 26 images. Moreover, PVT and ScalableViT are slower than Swin because the method that shrinks spatial tokens of Keys and Values may no longer be applicable in high-resolution images. Therefore, a Vision Transformer, friendly to high-resolution images and industrial scenes, needs to be developed with both higher accuracy and lower latency.

\section{Discussion and Conclusion}
\label{sec:majhead}

In this paper, we present the first publicly available Smartphone Screen Glass Dataset for defect detection.
We adapted professional capturing device and non-single workstations to collect images which involves various categories of defects commonly existing in actual producing procedure.
Then, we elaborately analysed the object distribution of this dataset
Next, abundant experiments are conducted to show the performance of popular methods on proposed dataset. 
Based on the comparison of experimental results, we find that the Vision Transformers perform worse and are much slower than their CNN counterparts at high resolution input. At the same time, a dynamic assignment strategy during training is very important to this dataset.
In the future, we will continue the investigation to develop more promising and approachable methods to improve the detection effect in the testing process. We hope this paper paves the way for the application of computer vision in the defect detection of screens.


\section{ACKNOWLEDGEMENT}
\label{sec:foot}
This work was supported by the National Key R\&D Program of China 505 (Grant No.2020AAA0108303), the Shenzhen Science and Technology Project (Grant No.JCYJ20200109143
041798) and Shenzhen Stable Supporting Program (WDZC202
00820200655001). Partial samples and analytical methods are provided by Shenzhen Zhihan Equipment Ltd., Li Xinghui and Wang Xiaohao.




\vfill\pagebreak

\bibliographystyle{IEEEbib}
\bibliography{strings,refs}

\begin{thebibliography}{10}

\bibitem{bao2021triplet}
Yanqi Bao, Kechen Song, Jie Liu, Yanyan Wang, Yunhui Yan, Han Yu, and Xingjie
  Li,
\newblock ``Triplet-graph reasoning network for few-shot metal generic surface
  defect segmentation,''
\newblock {\em IEEE TIM}, vol. 70, pp. 1--11, 2021.

\bibitem{song2013micro}
Kechen Song and Yunhui Yan,
\newblock ``Micro surface defect detection method for silicon steel strip based
  on saliency convex active contour model,''
\newblock {\em Math. Probl. Eng}, vol. 2013, 2013.

\bibitem{gan2017hierarchical}
Jinrui Gan, Qingyong Li, Jianzhu Wang, and Haomin Yu,
\newblock ``A hierarchical extractor-based visual rail surface inspection
  system,''
\newblock {\em IEEE Sens. J}, vol. 17, no. 23, pp. 7935--7944, 2017.

\bibitem{pallemulla2021defect}
PSH Pallemulla, SJ~Sooriyaarachchi, CR~de~Silva, and CD~Gamage,
\newblock ``Defect detection in woven fabrics by analysis of co-occurrence
  texture features as a function of gray-level quantization and window size,''
\newblock {\em ENGINEER}, vol. 54, no. 04, pp. 55--64, 2021.

\bibitem{pramerdorfer2015dataset}
Christopher Pramerdorfer and Martin Kampel,
\newblock ``A dataset for computer-vision-based pcb analysis,''
\newblock in {\em MVA}. IEEE, 2015.

\bibitem{Deitsch2021}
Sergiu Deitsch, Claudia Buerhop-Lutz, Evgenii Sovetkin, Ansgar Steland, Andreas
  Maier, Florian Gallwitz, and Christian Riess,
\newblock ``Segmentation of photovoltaic module cells in uncalibrated
  electroluminescence images,''
\newblock {\em Mach Vis Appl}, vol. 32, no. 4.

\bibitem{Faster}
Shaoqing Ren, Kaiming He, Ross~B. Girshick, and Jian Sun,
\newblock ``Faster {R-CNN:} towards real-time object detection with region
  proposal networks,''
\newblock {\em {IEEE} Trans. Pattern Anal. Mach. Intell.}, vol. 39, no. 6, pp.
  1137--1149, 2017.

\bibitem{FCOS}
Zhi Tian, Chunhua Shen, Hao Chen, and Tong He,
\newblock ``{FCOS:} fully convolutional one-stage object detection,''
\newblock in {\em ICCV}, 2019.

\bibitem{YOLOX}
Zheng Ge, Songtao Liu, Feng Wang, Zeming Li, and Jian Sun,
\newblock ``{YOLOX:} exceeding {YOLO} series in 2021,''
\newblock {\em arXiv:2107.08430}, 2021.

\bibitem{mmdetection}
Kai Chen, Jiaqi Wang, Jiangmiao Pang, Yuhang Cao, Yu~Xiong, Xiaoxiao Li,
  Shuyang Sun, Wansen Feng, Ziwei Liu, Jiarui Xu, Zheng Zhang, Dazhi Cheng,
  Chenchen Zhu, Tianheng Cheng, Qijie Zhao, Buyu Li, Xin Lu, Rui Zhu, Yue Wu,
  Jifeng Dai, Jingdong Wang, Jianping Shi, Wanli Ouyang, Chen~Change Loy, and
  Dahua Lin,
\newblock ``{MMDetection}: Open mmlab detection toolbox and benchmark,''
\newblock {\em arXiv:1906.07155}, 2019.

\bibitem{COCO}
Tsung{-}Yi Lin, Michael Maire, Serge~J. Belongie, James Hays, Pietro Perona,
  Deva Ramanan, Piotr Doll{\'{a}}r, and C.~Lawrence Zitnick,
\newblock ``Microsoft {COCO:} common objects in context,''
\newblock in {\em ECCV}, 2014.

\bibitem{ResNet}
Kaiming He, Xiangyu Zhang, Shaoqing Ren, and Jian Sun,
\newblock ``Deep residual learning for image recognition,''
\newblock in {\em CVPR}, 2016.

\bibitem{SGD}
Sebastian Ruder,
\newblock ``An overview of gradient descent optimization algorithms,''
\newblock {\em arXiv:1609.04747}, 2016.

\bibitem{Swin}
Ze~Liu, Yutong Lin, Yue Cao, Han Hu, Yixuan Wei, Zheng Zhang, Stephen Lin, and
  Baining Guo,
\newblock ``Swin transformer: Hierarchical vision transformer using shifted
  windows,''
\newblock in {\em ICCV}, 2021.

\bibitem{PVT}
Wenhai Wang, Enze Xie, Xiang Li, Deng{-}Ping Fan, Kaitao Song, Ding Liang, Tong
  Lu, Ping Luo, and Ling Shao,
\newblock ``Pyramid vision transformer: {A} versatile backbone for dense
  prediction without convolutions,''
\newblock in {\em ICCV}, 2021.

\bibitem{ScalableViT}
Rui Yang, Hailong Ma, Jie Wu, Yansong Tang, Xuefeng Xiao, Min Zheng, and Xiu
  Li,
\newblock ``Scalablevit: Rethinking the context-oriented generalization of
  vision transformer,''
\newblock {\em arXiv:2203.10790}, 2022.

\bibitem{UniFormer}
Kunchang Li, Yali Wang, Peng Gao, Guanglu Song, Yu~Liu, Hongsheng Li, and
  Yu~Qiao,
\newblock ``Uniformer: Unified transformer for efficient spatial-temporal
  representation learning,''
\newblock in {\em ICLR}, 2022.

\bibitem{AdamW}
Ilya Loshchilov and Frank Hutter,
\newblock ``Decoupled weight decay regularization,''
\newblock in {\em ICLR}, 2019.

\bibitem{Cascade}
Zhaowei Cai and Nuno Vasconcelos,
\newblock ``Cascade {R-CNN:} delving into high quality object detection,''
\newblock in {\em CVPR}, 2018.

\bibitem{RetinaNet}
Tsung{-}Yi Lin, Priya Goyal, Ross~B. Girshick, Kaiming He, and Piotr
  Doll{\'{a}}r,
\newblock ``Focal loss for dense object detection,''
\newblock in {\em ICCV}, 2017.

\bibitem{ATSS}
Shifeng Zhang, Cheng Chi, Yongqiang Yao, Zhen Lei, and Stan~Z. Li,
\newblock ``Bridging the gap between anchor-based and anchor-free detection via
  adaptive training sample selection,''
\newblock in {\em CVPR}, 2020.

\bibitem{GFL}
Xiang Li, Wenhai Wang, Lijun Wu, Shuo Chen, Xiaolin Hu, Jun Li, Jinhui Tang,
  and Jian Yang,
\newblock ``Generalized focal loss: Learning qualified and distributed bounding
  boxes for dense object detection,''
\newblock in {\em NeurIPS}, 2020.

\end{thebibliography}

\end{document}